\documentstyle[aclap]{article}
\title {Using Terminological Knowledge Representation Languages to Manage Linguistic Resources} 
\author{Pamela W.\ Jordan\\
Intelligent Systems Program\\
University of Pittsburgh\\
Pittsburgh PA 15260\\
jordan@isp.pitt.edu}

\begin{document}
\maketitle

\begin{abstract}

  I examine how terminological languages can be used to manage
linguistic data during NL research and development.  In particular,
I consider the lexical semantics task of characterizing semantic
verb classes and show how the language can be extended to flag
inconsistencies in verb class definitions,
identify the need for new verb classes, and identify appropriate
linguistic hypotheses for a new verb's behavior.
\end{abstract}

\section{Introduction}

Problems with consistency and completeness can arise when writing a
wide-coverage grammar or analyzing lexical data since both tasks
involve working with large amounts of data.  Since terminological
knowledge representation languages have been valuable for managing
data in other applications such as a software information system that
manages a large knowledge base of plans \cite{devanbu-litman:1991a},
it is worthwhile considering how these languages can be used in
linguistic data management tasks.  In addition to inheritance,
terminological systems provide a criterial semantics for links and
automatic classification which inserts a new concept into a taxonomy
so that it directly links to concepts more general than it and more
specific than it \cite{woods-schmolze:1992a}.

Terminological languages have been used in NLP applications
for lexical representation
\cite{burkert}, and grammar representation
\cite{brachman-schmolze:1991a}, and to assist in the
acquisition and maintenance of domain specific lexical semantics
knowledge \cite{ayuso-etal:1987}.  Here I explore additional
linguistic data management tasks.  In particular I examine how a
terminological language such as Classic \cite{brachman-etal:1991a} can
assist a lexical semanticist with the management of verb classes.  In
conclusion, I discuss ways in which terminological languages can
be used during grammar writing.

Consider the tasks that confront a lexical
semanticist.  The regular participation of verbs belonging to a
particular semantic class in a limited number of syntactic
alternations is crucial in lexical semantics.  A popular research
direction assumes that the syntactic behavior of a verb is
systematically influenced by its meaning \cite{Levin,HaleKeyser} and
that any set of verbs whose members pattern together with respect to
syntactic alternations should form a semantically coherent class
\cite{Levin}.  Once such a class is identified, the meaning component
that the member verbs share can be identified.  This gives
further insight into lexical representation for the words in the
class \cite{Levin}.

Terminological languages can support three important functions in this
domain.  First, the process of representing the system in a taxonomic
logic can serve as a check on the rigor and precision of the original
account.  Once the account is represented, the terminological
system can flag inconsistencies.  Second, the classifier can identify
an existing verb class that might explain an unassigned verb's
behavior.  That is, given a set of syntactically analyzed sentences
that exemplify the syntactic alternations allowed and disallowed for
that verb, the classifier will provide appropriate linguistic
hypotheses.  Third, the classifier can identify the need for new verb
classes by flagging verbs that are not members of any existing,
defined verb classes.  Together, these functions provide tools
for the lexical semanticist that are potentially very useful.

The second and third of these three functions can be provided in two
steps: (1) classifying each alternation for a particular verb
according to the type of semantic mapping allowed for the verb and its
arguments; and (2) either identifying the verb class that has the
given pattern of classified alternations or using the pattern to form
the definition of a new verb class.

\section{Sentence Classification}

The usual practice in investigating the alternation patterning of a
verb is to construct example sentences in which simple, illustrative
noun phrases are used as arguments of a verb.  The sentences in (1)
exemplify two familiar alternations of {\it give}.

\begin{small}
 \begin{enumerate}
 \item[(1)]a. John gave Mary a book
 \item[ ]b. John gave a book to Mary.
 \end{enumerate}
\end{small}

 Such sentences exemplify an alternation that belongs to the
alternation pattern of their verb.\footnote{In the examples that I
will consider, and in most examples used by linguists to test
alternation patterns, there will only be one verb; this is the verb
to be tested.} I will call this the {\em alternation type} of the
test sentence.

To determine the alternation type of a test sentence, the sentence must be
syntactically analyzed so that its grammatical functions
(e.g. subject, object) are marked.  Then, given semantic feature
information about the words filling those grammatical functions (GFs),
and information about the possible argument structures for the verb in
the sentence and the semantic feature restrictions on these arguments,
it is possible to find the argument structures appropriate to the
input sentence.  Consider the sentences and
descriptions shown below for $pour$:

\begin{small}
\begin{enumerate}
\item[(2)]a. [Mary$_{subj}$] poured [Tina$_{obj}$] [a glass of milk$_{io}$].
 \item[]b. [Mary$_{subj}$] poured [a glass of milk$_{obj}$] for [Tina$_{ppo}$].
\end{enumerate}

\begin{tabular}{ll}
$pour_{1}$: &subj $\rightarrow$ agent$_{[volitional]}$ \\ 
&obj $\rightarrow$ recipient$_{[volitional]}$ \\ 
&io $\rightarrow$ patient$_{[liquid]}$ \\
$pour_{2}$: &subj $\rightarrow$ agent$_{[volitional]}$ \\ 
&obj $\rightarrow$ patient$_{[liquid]}$ \\ 
&ppo $\rightarrow$  recipient$_{[volitional]}$\\
\end{tabular}
\end{small}

Given the semantic type restrictions and the GFs, $pour_{1}$ describes
(2a) and $pour_{2}$, (2b).  The mapping from the GFs to the
appropriate argument structure is similar to lexical rules in the LFG
syntactic theory except that here I semantically type the arguments.
To indicate the alternation types for these sentences, I call sentence
(2a) a benefactive-ditransitive and sentence (2b) a
benefactive-transitive.

Classifying a sentence by its alternation type requires linguistic and
world knowledge.  World knowledge is used in the definitions of nouns
and verbs in the lexicon and describes high-level entities, such as
events, and animate and inanimate objects.  Properties (such as {\sc
liquid}) are used to define specialized entities. For example, the
property {\sc non-consumable} ({\sc small capitals} indicate Classic
concepts in my implementation) specializes a {\sc liquid-entity} to
define {\sc paint} and distinguish it from {\sc water}, which has the
property that it is {\sc consumable}.  Specialized {\sc event}
entities are used in the definition of verbs in the lexicon and
represent the argument structures for the verbs.

The linguistic knowledge needed to support sentence classification
includes the definitions of (1) verb types such as intransitive,
transitive and ditransitive; (2) verb definitions; and (3) concepts
that define the links between the GFs and verb argument
structures as represented by events.

Verb types ({\sc subcategorization}s) are defined according to the GFs
found in the sentence.  For example, (2a) classifies as {\sc
ditransitive} and (2b) as a specialized {\sc transitive} with a PP.
Once the verb type is identified, verb definitions ({\sc verb}s) are
needed to provide the argument structures.  A {\sc verb} can have
multiple senses which are instances of {\sc event}s, for example the
verb ``pour'' can have the senses $pour$ or $prepare$, with the
required arguments shown below.\footnote{For generality in the
implementation, I use arg$_{1} \ldots$ arg$_{n}$ for all event
definitions instead of agent $\ldots$ patient or preparer $\ldots$
preparee.} Note that $pour_{1}$ and $pour_{2}$ in (2) are
subcategorizations of $prepare$.

\begin{small}
\begin{tabular}{ll}
$pour$: &pourer$_{[volitional]}$ \\ 
        &pouree$_{[inanimate\--container]}$ \\
        &poured$_{[inanimate\--substance]}$ \\
$prepare$: &preparer$_{[volitional]}$ \\ 
           &preparee$_{[liquid]}$ \\ 
           &prepared$_{[volitional]}$
\end{tabular}
\end{small}

For a sentence to classify as a particular {\sc alternation}, a legal
linking must exist between an {\sc event} and the {\sc
subcategorization}.  Linking involves restricting the fillers of the
GFs in the {\sc subcategorization} to be the same as the arguments in
an {\sc event}.  In Classic, the {\tt same-as} restriction is limited
so that either both attributes must be filled already with the same
instance or the concept must already be known as a {\sc
legal-linking}.  Because of this I created a test (written in
LISP) to identify a {\sc legal-linking}.  The test inputs are the
sentence predicate and GF fillers arranged in the order of the event
arguments against which they are to be tested.  A linking is legal
when at least one of the events associated with the verb can be linked
in the indicated way, and all the required arguments are filled.

Once a sentence passes the linking test, and classifies as a
particular {\sc alternation}, a rule associated with the {\sc
alternation} classifies it as a specialization of the concept.  This
causes the {\sc event} arguments to be filled with the appropriate GF
fillers from the {\sc subcategorization}.  A side-effect of the
alternation classification is that the {\sc event} classifies as a
specialized {\sc event} and indicates which sense of the verb is used
in the sentence.

\section{Semantic Class Classification} \label{verb}

The semantic class of the verb can be identified once the example sentences
are classified by their alternation
type.  Specialized {\sc verb-class}es are defined by their good and
bad alternations.  Note that {\sc verb} defines one verb whereas 
{\sc verb-class} describes a set of verbs (e.g. spray/load class).
Which {\sc alternation}s are associated with a {\sc verb-class} is a
matter of linguistic evidence; the linguist discovers these
associations by testing examples for grammaticality.  To assist in
this task, I provide two tests, {\tt have-instances-of} and {\tt
have-no-instances-of}.  The {\tt have-instances-of} test for an {\sc
alternation} searches a corpus of good sentences or bad sentences and
tests whether at least one instance of the specified {\sc
alternation}, for example a benefactive-ditransitive, is present.

A bad sentence with all the required verb arguments will classify as
an {\sc alternation} despite the ungrammatical syntactic realization,
while a bad sentence with missing required arguments will only classify
as a {\sc subcategorization}.  The {\tt have-no-instances-of} test for
a {\sc subcategorization} searches a corpus of bad sentences and tests
whether at least one instance of the specified {\sc
subcategorization}, for example {\sc transitive}, is present as the most
specific classification.

\section{Discussion}

The ultimate test of this approach is in how well it will scale up.
The linguist may choose to add knowledge as it is needed
or may prefer to do this work in batches.  To support the batch
approach, it may be useful to extract detailed subcategorization
information from English learner's dictionaries.  Also it will be
necessary to decide what semantic features are needed
to restrict the fillers of the argument structures.  
Finally, there is the problem of collecting complete sets of example
sentences for a verb.  In general, a corpus of tagged
sentences is inadequate since it rarely includes negative
examples and is not guaranteed to exhibit the full range of
alternations.  In applications where a domain specific corpus
is available (e.g. the Kant MT project \cite{Mitamura:etal:93}), the
full range of relevant alternations is more likely.  However, the lack
of negative examples still poses a problem and would require the
project linguist to create appropriate negative examples or manually
adjust the class definitions for further differentiation.

While I have focused on a lexical research tool, an area I will
explore in future work is how classification could be used in grammar
writing.  One task for which a terminological language is appropriate
is flagging inconsistent rules.  When writing and maintaining a
large grammar, inconsistent rules is one type of grammar writing bug that occurs.
For example, the following three rules are inconsistent since feature$_{1}$
of NP and feature$_{1}$ of VP would not unify in rule 1 given the values
assigned in 2 and 3.

\begin {small}
\begin{tabular}{lll}
1)&S $\rightarrow$ NP VP\\
&$<$NP feature$_{1}>$ = $<$VP feature$_{1}>$\\
2)& NP $\rightarrow$ det N\\
&$<$N feature$_{1}>$ = +\\
&$<$NP$>$ = $<$N$>$\\
3)&VP $\rightarrow$ V\\
&$<$V feature$_{1}>$ = $-$\\
&$<$VP$>$ = $<$V$>$
\end{tabular}
\end{small}

\section {Conclusion}

I have shown how a terminological language, such as Classic, can be
used to manage lexical semantics data during analysis with two minor
extensions.  First, a test to identify {\sc legal-linking}s is
necessary since this cannot be directly expressed in the language and
second, set membership tests, {\tt have-instances-of} and {\tt
have-no-instances-of} are necessary since this type of expressiveness
is not provided in Classic.  While the solution of several knowledge 
acquisition issues would result in a friendlier
tool for a linguistics researcher, the tool still performs a useful
function.

\end{document}